\documentclass{article} 
\usepackage{iclr2024_conference,times}

\usepackage{amsmath,amsfonts,bm}

\def\eqref#1{equation~\ref{#1}}

\def\1{\bm{1}}

\def\vc{{\bm{c}}}

\def\vx{{\bm{x}}}

\DeclareMathAlphabet{\mathsfit}{\encodingdefault}{\sfdefault}{m}{sl}
\SetMathAlphabet{\mathsfit}{bold}{\encodingdefault}{\sfdefault}{bx}{n}

\usepackage{hyperref}
\usepackage{url}
\usepackage{graphicx}
\usepackage{pgfplots}
\usepackage{tikzscale}
\usepackage{wrapfig}
\usepackage{subfig}
\usepackage{cleveref}
\usepgfplotslibrary{groupplots}
\usepackage{subcaption}

\title{SUPClust: Active Learning at the Boundaries}

\author{Yuta Ono, Till Aczel, Benjamin Estermann \& Roger Wattenhofer\\
ETH Zürich\\
\texttt{\{yutono,taczel,estermann,wattenhofer\}@ethz.ch} \\
}

\iclrfinalcopy 
\begin{document}

\maketitle

\vspace{-1pt}
\begin{abstract}
    Active learning is a machine learning paradigm designed to optimize model performance in a setting where labeled data is expensive to acquire.
    In this work, we propose a novel active learning method called SUPClust that seeks to identify points at the decision boundary between classes.
    By targeting these points, SUPClust aims to gather information that is most informative for refining the model's prediction of complex decision regions.
    We demonstrate experimentally that labeling these points leads to strong model performance. 
    This improvement is observed even in scenarios characterized by strong class imbalance.
\end{abstract}

\section{Introduction}
    Progress in deep learning for classification tasks has been following an impressive pace in recent years \citep{Ioffe2015Batch, Dosovitskiy2021ViT, Srivastava_2024_WACV}.
    In order to achieve high classification accuracy on a target dataset, many of these methods necessitate a substantial amount of annotated data.
    However, in many settings, annotating data is both time-consuming and costly, posing a challenge to the application of these successful methods in scenarios with limited resources.
    One of the ways to mitigate this problem is active learning.
    Active learning aims to maximize performance by selecting the most informative and valuable data points to be annotated for model training.

    \begin{wrapfigure}{r}{0.4\textwidth}
        \centering
        \vspace{-20pt}
        \includegraphics[width = 0.4 \textwidth]{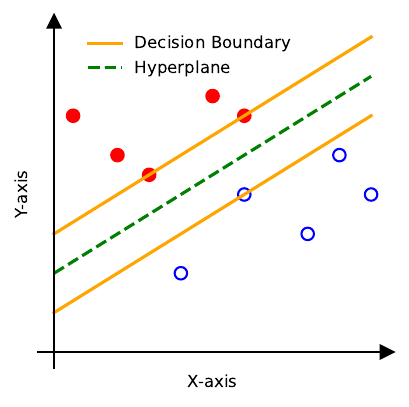}
        \caption{Decision boundary of an SVM classifier.
        }
        \label{fig:decision_boundary}
    \end{wrapfigure}

    But how can a model correctly classify points of different classes?
    Classical support vector machines (SVMs) search for a hyperplane that separates two classes with the largest possible margin (see \Cref{fig:decision_boundary}). 
    The points that lie on this decision boundary are called \textit{support vectors}.
    In other words, these support vectors define the boundary of all samples of a class and are critical for a model to know in order to correctly separate the classes.
    We hypothesize that points close to the decision boundary are similarly relevant for neural network-based models.

    In this work, we propose a novel active learning method (SUPClust) that tries to identify these points so that they can be annotated.
    Since the labels of the points are not known a priori, we rely on self-supervised representation learning in combination with clustering in order to break down the high-dimensional input space.
    For each cluster, we then identify the points close to a neighboring cluster, thereby selecting potential support vector points.
    Thanks to selecting points from all clusters, we ensure a broad coverage of the input space.
    In practice, data distributions often include outliers and the decision boundary between different classes is not always clearcut.
    For this reason, we further constrain our points to be somewhat typical according to a typicality metric introduced by \citet{Hacohen2022-lr}.

    Our experimental evaluation demonstrates the merit of sampling points closer to the decision boundary, underscored by the strong performance compared to baseline active learning methods.
    SUPClust manages to not only mitigate the ``cold start problem~\citep{Mittal2019-oh}'', it also shows strong performance in datasets with strong class imbalance.
    In ablation experiments, we ensure that all building blocks of SUPClust are necessary and contribute to the final result.

\section{Related Work}
    Various active learning methods have been proposed to this end, which can be categorized in uncertainty-based and diversity-based.
    Uncertainty-based approaches \citep{Lewis1994Sequential, Joshi2009Multi, Gal2017-fg} leverage the prediction uncertainty of the classification model under training to select informative data samples for annotation.
    Diversity-based approaches \citep{Sener2017-hm, Yehuda2022-yv, Hacohen2022-lr} aim to annotate a diverse range of samples spanning the complete data distribution, avoiding the selection of too similar ones.
    There also exist hybrid methods \citep{Ash2019-cn} which try to identify samples that have high uncertainty and are diverse at the same time.
    Some of these models rely on embeddings learned during self-supervised pre-training.
    Self-supervised learning involves training a model on a pretext task, allowing it to learn valuable representations without relying on explicit external labels. 
    These representations complement the active learning task because they contain important information about the structure of the data distribution.

    We give a short summary of the most used uncertainty-based approaches.
    Least confidence \citep{Lewis1994Sequential}, Entropy \citep{Joshi2009Multi}, and Margin all select uncertain samples according to an uncertainty measure based on the output logits of the trained classifier.
    DBAL and BALD \citep{Gal2017-fg} on the other hand utilize Bayesian convolutional neural networks as a classifier and then select samples based on the highest entropy in the classifier or largest information gain.
    Many of these methods suffer from the ``cold start problem'', where their performance in low-budget regimes is worse than randomly selecting samples.
    This is possibly caused by the uncertainty estimates to be bad when the underlying model is not trained sufficiently due to limited labeled samples.
    SUPClust avoids this issue by selecting samples close to the decision border between clusters in the embedding space of a self-supervised pre-trained model.

    In the realm of diversity-based methods, Coreset \citep{Sener2017-hm} queries diverse samples through the selection of points that form a minimum radius cover of the remaining samples in the unlabeled pool. 
    To do this, Coreset works on the embeddings generated by the penultimate layer of the classifier.
    In comparison, ProbCover \citep{Yehuda2022-yv} and TypiClust \citep{Hacohen2022-lr} rely on the embeddings of a self-supervised pre-trained model.
    ProbCover selects a maximum cover set for fixed-sized balls in this pre-trained embedding space.
    Typiclust builds clusters in the embedding space.
    From each cluster, it then selects the most typical sample.
    This combination ensures both broad coverage of the input space as well as selecting informative points, which shows in its strong performance in low-budget regimes.
    Typicality is measured in the following way:
    \begin{equation}
      Typicality(\vx) = \left(\frac{1}{K}\sum_{\vx_n\in K-\mathrm{NN}(\vx)}\|\vx-\vx_n\|\right)^{-1} \label{eq:def-typicality}
    \end{equation}
    Here, \(K-\mathrm{NN}(\vx)\) is a set of \(K\) nearest neighbors of \(\vx\) in an embedding space.
    SUPClust also relies on typicality in order to ensure that the selected points are still representative of the cluster they come from.

\section{SUPClust}
    \begin{figure}
    \centering
    \begin{minipage}{.35\textwidth}
      \centering
      \vspace{1.24cm}
      \includegraphics[width=\linewidth]{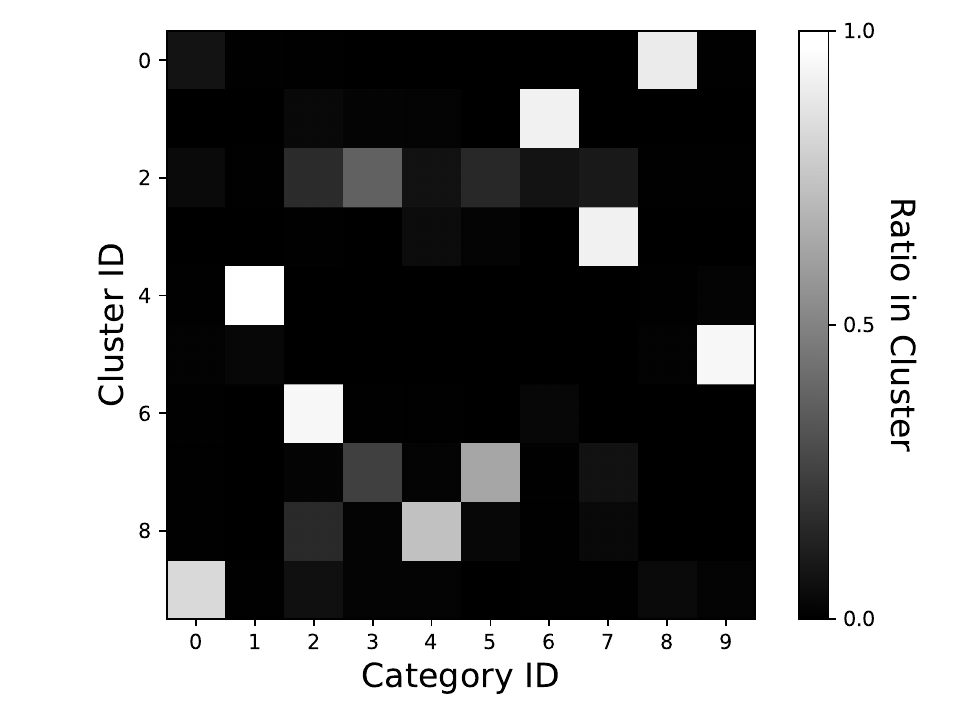}
      \captionof{figure}{Distribution of classes within each cluster on SimCLR embeddings for CIFAR-10. Cluster boundaries align with category boundaries.}
      \label{fig:clustering}
    \end{minipage} 
    \hfill 
    \begin{minipage}{.62\textwidth}
      \centering
      \begin{minipage}{0.48\textwidth}
            \centering
            \caption*{TypiClust}
        \end{minipage}
        \hfill
        \begin{minipage}{0.48\textwidth}
            \centering
            \caption*{SUPClust}
        \end{minipage}
      \includegraphics[width=.48\linewidth]{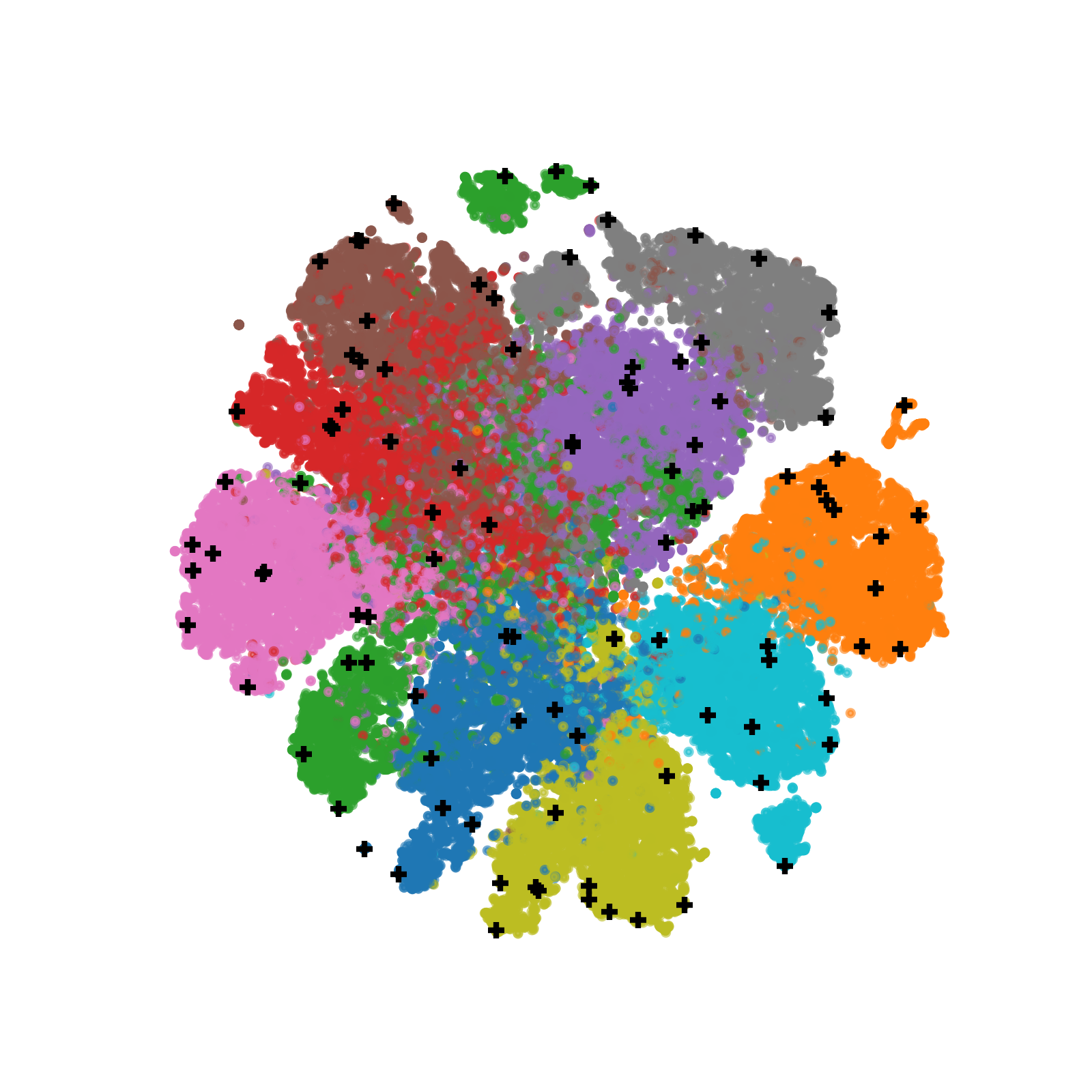}
      \includegraphics[width=.48\linewidth]{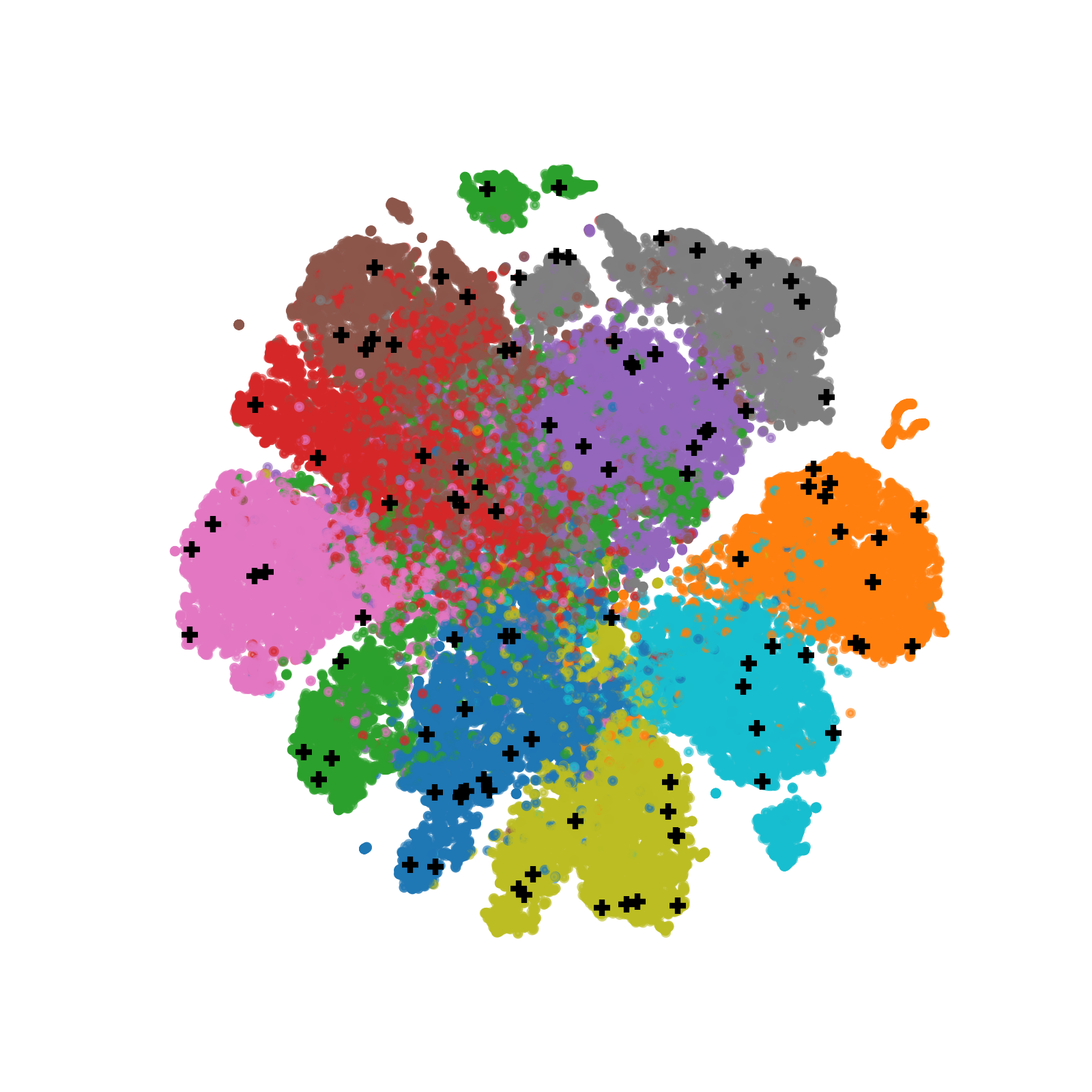}
      \captionof{figure}{t-SNE plots of 100 queried instances by TypiClust and SUPClust (ours) in the CIFAR-10 embedding space. Colors represent the categories.
        For clusters on the ``edge'' of the data distribution, SUPClust tends to select samples that are closer to other clusters in the embedding space.}
      \label{fig:tsne-comparison}
    \end{minipage}
    \end{figure}

    SVM classifiers are defined by a few key points located at the decision boundary between the categories. 
    Our querying strategy selects instances situated near the decision boundary, as they provide a strong signal to the learning process of neural network based models too.
    Traditional active learning methods have approached this problem by using model uncertainty as an indication for samples at the decision boundary.
    However, these methods suffer from the cold-start problem, where in low-budget scenarios, the model uncertainty is unable to identify hard instances.
    In this work, we introduce a novel method to find such samples by exploiting pre-trained representations. 
    We can see in Figure \ref{fig:clustering} on the example of CIFAR-10 that similar categories are clustered together in the representation space.
    As category boundaries align with cluster boundaries, we use clustering to identify samples of interest.
    To quantify proximity to the decision boundary, we compute, for each sample, the weighted mean distance to all other cluster centers.
    The weights are the same for all samples within a cluster and are dependent on the distance of the cluster center to all other cluster centers.
    Clusters positioned at the ``edge'' of the data distribution select an instance that is close to the nearest cluster.
    Conversely, clusters in the ``middle'' of the distribution do not select instances that are close to just one of the clusters.
    To normalize the weights to 1, we use the softmax function with the negative L2 distance as the logits and the temperature parameter $T$.
    For a point in cluster $i$, the weight to the cluster $j$ is given by
    \newcommand{\norm}[1]{\left\lVert#1\right\rVert}
    \begin{equation}
        w_i^j = \frac{\exp \left(-\frac{\norm{\vc_i-\vc_j}}{T}\right)}{\sum_{k\in C\setminus \{i\}} \exp \left(-\frac{\norm{\vc_i-\vc_k}}{T}\right)},
    \end{equation}
    where $\vc_i$, $\vc_j$ and $\vc_k$ are the centers of cluster $i$, $j$ and $k$ respectively, and $C$ is the set of all clusters.
    For cluster $i$, we select the sample $\vx$, that has the minimum distance, or maximum $SUP$ to the decision boundary computed by Equation \ref{equ:distance}.
    \begin{equation}
        SUP(\vx) = \left(\sum_{j\in C\setminus \{i\}} w_i^j \norm{\vx-\vc_j}\right)^{-1}
        \label{equ:distance}
    \end{equation}

\begin{figure}[!htb]
        \centering
        \includegraphics[width=\linewidth]{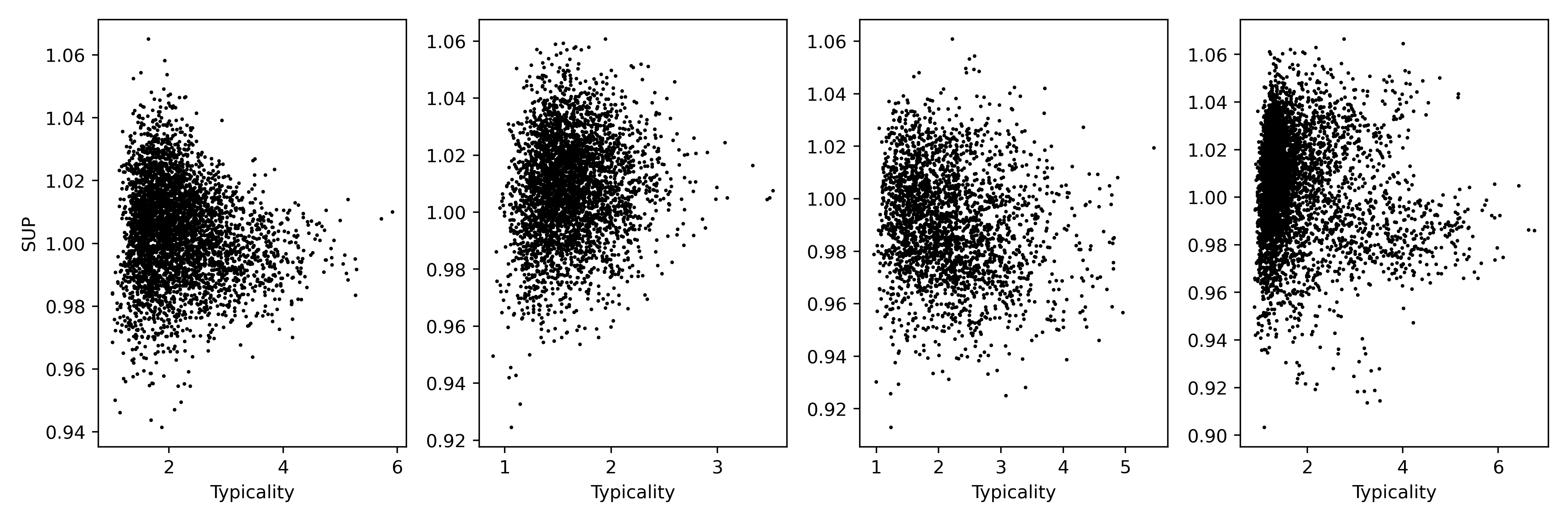}
        \caption{Relationship between typicality and SUP on CIFAR-10 on 4 randomly selected clusters, with temperature $1$.     
        Typicality and SUP have no strong correlation, using both metrics to select instances can improve the querying strategy.
        }
        \label{fig:typicality-and-SUP}
    \end{figure}

    Real data distributions are noisy, contain outliers, and can not be separated by a hyperplane, thus sampling only based on $SUP$ leads to subpar performance.
    To avoid outliers on the decision boundary we combine typicality with $SUP$.
    Typicality and $SUP$ are not correlated, see Figure \ref{fig:typicality-and-SUP}, thus using both metrics for sample selection can improve performance.
    
    Our proposed strategy \textbf{SUPClust} consists of 4 parts. 
    1) Train a self-supervised model on the unlabeled pool.
    2) Partition the data into $N$ clusters, where $N$ is the number of labeled samples after the end of the current step. 
    Excluding clusters that contain samples from the already labeled pool, selecting as many of the biggest clusters as samples are queried.
    3) In each cluster filter the top 10\% of samples based on typicality. 
    4) Select the sample with the highest $SUP$. 
    In Figure \ref{fig:tsne-comparison} we compare samples queried by TypiClust and and SUPClust. 
    Notably, TypiClust chooses more samples from the ``edge'' of the data distribution, whereas SUPClust prioritizes samples that lie closer to other categories.

\section{Results}
\subsection{Experimental setup}

    All strategies are evaluated on image classification tasks using CIFAR-10, CIFAR-100~\citep{krizhevsky2009learning}, CIFAR-10-LT~\citep{Cao2019-zn}, and ISIC-2019 \citep{Kassem2020-cg}. 
    CIFAR-10 and CIFAR-100 consist of 60k natural images of size 32x32 with 10/100 classes. 
    CIFAR-10-LT is a class-imbalanced subset of CIFAR-10. 
    We apply an imbalanced factor of 50, meaning a 50-fold difference in the number of images between the most and least frequent class.
    ISIC-2019 consists of 25331 skin cancer images with 8 imbalanced classes.
    To standardize the image dimensions, all images are resized to 224x224 pixels.
    In alignment with TypiClust, we adopt tiny and small budget sizes, involving querying step sizes 1 and 5 times the number of classes respectively.
    
    We evaluate AL strategies in the following two frameworks. 
    1) Fully supervised (FSL): training a deep neural network, ResNet18~\citep{He2015-wd}, exclusively on the labeled set which is acquired by active queries.
    2) Fully supervised with self-supervised embedding (SSL): training a linear classifier on the labeled embeddings obtained by active queries. 
    These self-supervised embeddings for the classifier are obtained from a pre-trained SimCLR \citep{Chen2020Simple}.
    Within these frameworks, we compare SUPClust to nine baseline strategies: Random, Margin, Least confidence, Entropy, BALD, Coreset, DBAL, TypiClust, ProbCover.
    For the clustering and sampling with TypiClust and SUPClust, we use SimCLR representations, namely the ResNet18 backbone for CIFAR-10, CIFAR-10-LT50 and ISIC-2019, and the ResNet34 for CIFAR-100.
   
\subsection{Ablation study}

    \begin{wrapfigure}{r}{0.45\textwidth}
        \centering
        \vspace{-25pt}
        \includegraphics[width=0.45\textwidth]{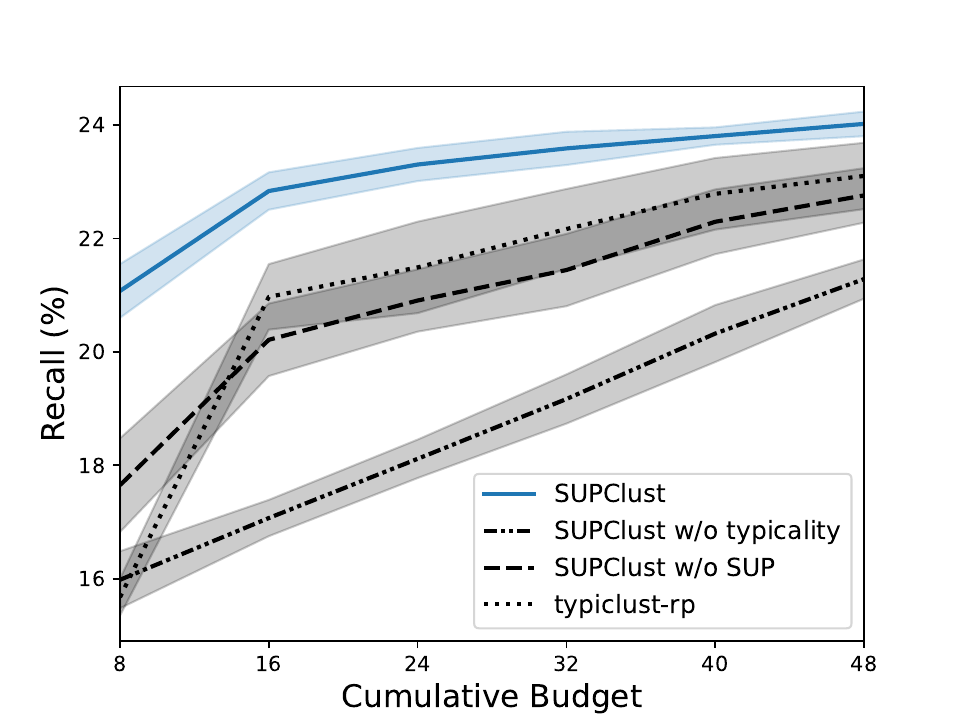}
        \caption{Ablation study on ISIC-2019 with budget=8 and with self-supervised embeddings}
        \label{fig:ablation-results}
    \end{wrapfigure}
    To assess the significance of individual components within SUPClust, we conduct ablation experiments for each component.
    We display the results in \Cref{fig:ablation-results}.
    When leaving out our SUP-based acquisition metric (SUPClust w/o SUP) and instead selecting a sample randomly from the top 10\% typical samples within each cluster, the performance noticeably declines, falling below that of TypiClust.
    Similarly, relying solely on SUP without considering typicality for sample selection (SUPClust w/o typicality) fails to achieve the performance levels observed with other querying strategies.
    As a comparison, we also show the default TypiClust (typiclust-rp), which always selects the most typical sample of a cluster.
    Our results showcase that all components of SUPClust are necessary and contribute to its performance.

\subsection{Main results}
    \begin{figure}[!htb]
        \centering
        \includegraphics[width=\linewidth]{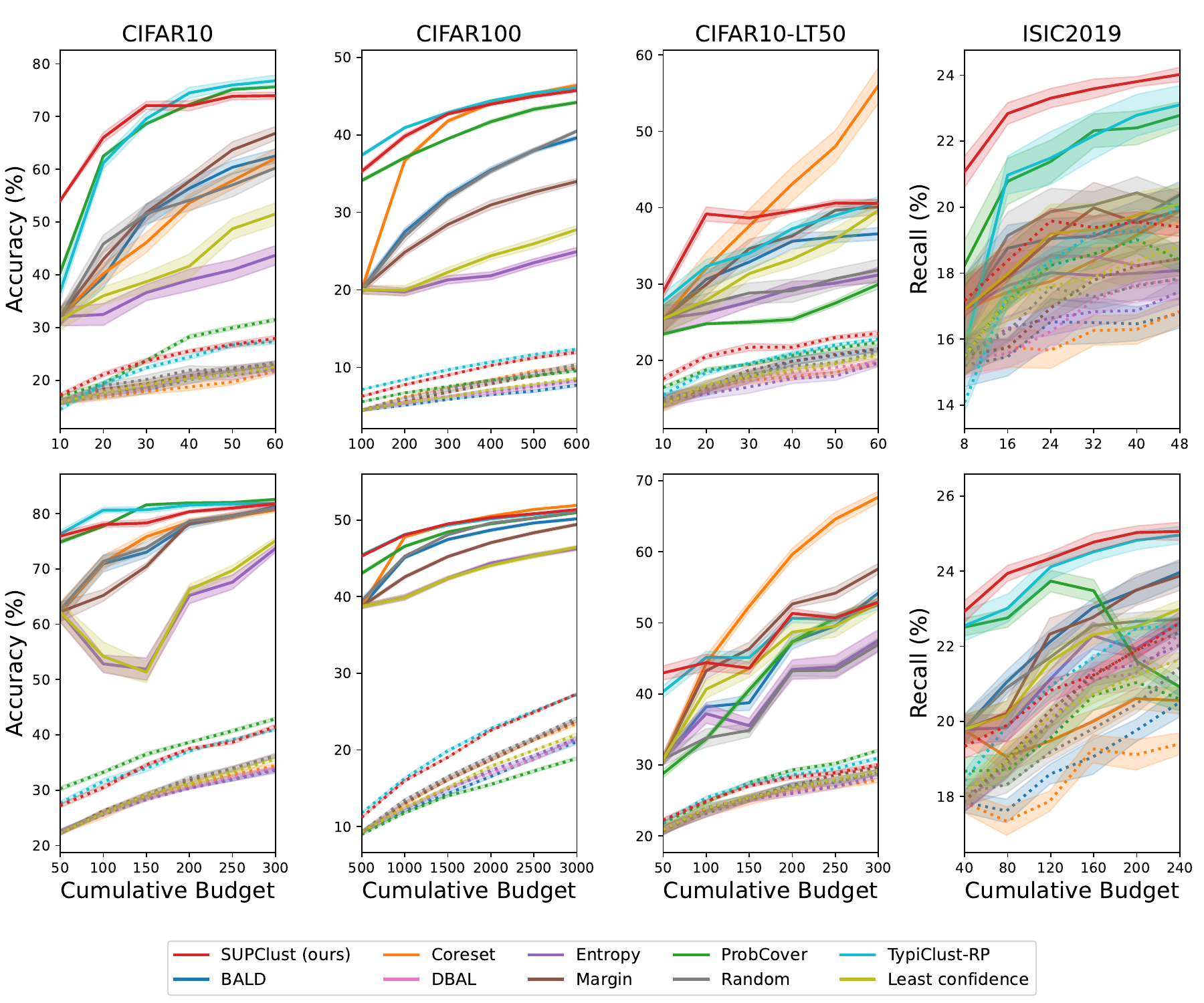}
        \caption{Results in the tiny (top) and small (bottom) budget regime. Solid lines represent results with the SSL setting, and dotted lines represent results with the FSL setting. The mean and the standard error with 10 different random seeds are shown. Our method (SUPClust) shows robust performance compared to other baselines, across all datasets and both data regimes.}
        \label{fig:results}
    \end{figure}

    We present the main results of our evaluation in \Cref{fig:results} for the tiny and the small budget regime.
    We can see that SUPClust performs well on all evaluated datasets, especially in imbalanced settings.
    On CIFAR10-LT50 and ISIC2019, SUPClust demonstrates a strong performance gain compared to TypiClust.
    We hypothesize that by selecting points according to maximum $SUP$, SUPClust is able to select more informative points relevant for distinguishing the classes, irrelevant of imbalance.
    In our low-budget regimes, diversity-based methods such as TypiClust, Coreset and ProbCover generally perform better than their uncertainty-based counterparts.
    This is to be expected, as uncertainty-based methods bring stronger benefits only in higher budget regimes.
    Building on the self-supervised pre-trained embeddings improves performance across all datasets.
    The performance of Coreset on CIFAR10-LT50 in the SSL setting is surprising.
    The embeddings of the pre-trained backbone allow Coreset to select very informative samples. 
    Unfortunately, when training in the FSL setting or on any other dataset, the performance of Coreset is diminished compared to other algorithms.

\section{Discussion}
    Active learning can bring performance benefits to settings where acquiring labeled data is expensive.
    Samples close to the decision boundary between categories provide a strong training signal. 
    The introduction of the novel $SUP$ metric provides a non-label-based means of quantifying the distance of each sample to the decision boundary. 
    Utilizing $SUP$, when selecting which samples to label for classifier training improves sample efficiency, especially in the low data budget regime.
    Our findings contribute to the broader understanding of active learning dynamics, shedding light on the relationship between the $SUP$ metric, typicality, and diversity.
    Exploring the changing relationship between diversity, typicality and the $SUP$ metric across various data regimes remains future work.

\newpage

\bibliography{iclr2024_conference}
\bibliographystyle{iclr2024_conference}

\end{document}